\documentclass[runningheads]{llncs}
\usepackage{caption}
\usepackage{subcaption}
\usepackage{graphicx}
\usepackage{cite}
\usepackage{amsmath,amssymb,amsfonts}
\usepackage{graphicx}
\usepackage{textcomp}
\usepackage{xcolor}
\usepackage{placeins}
\usepackage{pgfplots}
\begin{document}
\title{`Thy algorithm shalt not bear false witness':\\An Evaluation of Multiclass Debiasing Methods on Word Embeddings %\thanks{Supported by organization x.}
}
\titlerunning{An Evaluation of Multiclass Debiasing Methods on Word Embeddings}
\author{Thalea Schlender\and
Gerasimos Spanakis\orcidID{0000-0002-0799-0241}}
\institute{
Department of Data Science and Knowledge Engineering\\
Maastricht University\\
Maastricht, Netherlands
}
\maketitle              % typeset the header of the contribution
% 150 - 250 words
\begin{abstract}
With the vast development and employment of artificial intelligence applications, research into the fairness of these algorithms has been increased. Specifically, in the natural language processing domain, it has been shown that social biases persist in word embeddings and are thus in danger of amplifying these biases when used. As an example of social bias, religious biases are shown to persist in word embeddings and the need for its removal is highlighted. This paper investigates the state-of-the-art multiclass debiasing techniques: Hard debiasing, SoftWEAT debiasing and Conceptor debiasing. It evaluates their performance when removing religious bias on a common basis by quantifying bias removal via the Word Embedding Association Test (WEAT), Mean Average Cosine Similarity (MAC) and the Relative Negative Sentiment Bias (RNSB). By investigating the religious bias removal on three widely used word embeddings, namely: Word2Vec, GloVe, and ConceptNet, it is shown that the preferred method is ConceptorDebiasing. Specifically, this technique manages to decrease the measured religious bias on average by 82,42\%, 96,78\% and 54,76\% for the three word embedding sets respectively.

\end{abstract}
\section{Introduction}
\label{intro}
%introduce biases in AI
In recent years, there have been rapid advances in artificial intelligence and the accompanying vast development of machine learning applications. With the increased wide spread (commercial) employment of such applications it has become increasingly more vital to ensure their transparency, fairness and equality. Recent investigations of various application domains have shown that many of these applications exhibit several social biases endangering their fairness \cite{osoba2017intelligence}. Social biases describe the discrimination of certain identity groups based on, for example, their gender, race or religion. When social biases persist in machine learning applications, they run the danger of amplifying these biases. For instance, regarding social bias against minority groupds, it was found that these were recognized considerably less \cite{howard2018ugly}. To illustrate the real world consequences which minority group members face through biased algorithms, consider the use of these face /voice applications in sensitive areas such as medical diagnosis or the justice system. 
In cases like these, "the use of biased information could entail an extended and undeserved period of incarceration, which unjustly affects those who are arrested and possibly ruins the lives of their families" (p.7, \cite{howard2018ugly}). With respect to a medical application, "consider a revolutionary test for skin cancer that does not work on African Americans" (p.1, \cite{nelson2019bias}). 

%why biases occur in techno/social applications
Biases inherent in our society are, thus, perpetuated in the machine learning models, recorded by the model’s outcomes and, hence, threaten to treat various groups differently. To rectify the unequal treatment, the origin of biases in artificial intelligence needs to be examined and, consequently, removed. These biases in data driven applications may have myriad causes. One cause is the gathering of the data that is primarily done or planned by humans, which causes the data to be subject to similar biases as humans have. Moreover, the gathering process favours easy accessible and quantifiable data \cite{ntoutsi2020bias}, which may favour certain societal groups over others. Further, biases are captured in the under- / over-representation of societal groups in the dataset, which makes the complete data not representative of the end users anymore  \cite{ntoutsi2020bias}. Another origin of bias is data directly containing sensitive attributes, such as race or religion, or any proxy features for these. These proxy features may be well hidden, for instance a societal group may be represented in the post codes of communities. With the encoding of sensitive information, an algorithm can learn wrong causal inferences concerning these which can be hard to identify \cite{ntoutsi2020bias}.

%Introduce NLP social biases as a subproblem of this paper

The origins of bias mentioned above can be present in many representations of data. To provide an elaborate analysis, this paper will henceforth tend to textual data solely. To process textual data for an application, the data must be represented numerically. This is done via word embeddings, which attempt to capture the meaning and semantic relationships of a word and translate these to a real valued vector. Since word embeddings are learnt from possibly biased data, word embeddings themselves may contain biases, which could ripple through an application. Having outlined why the mitigation of these biases is vital and having introduced the domain of biased word embeddings, this paper will review work on analysis and mitigation of biased word embeddings, before presenting and evaluating various state-of-the-art post processing approaches to the mitigation of the found biases. Specifically, the attempted removal of multiclass social biases in three word embeddings is quantified on geometrical as well as on downstream evaluation metrics.

In order to highlight the results, the problem of religious bias is taken as a novel example for multi-class social bias. By doing so this paper aims to answer following research questions: 
\begin{itemize}
    \item To what extent are Religious biases, as an example for social bias, present in widely used word embeddings?
    \item How do state-of-the-art multiclass debiasing techniques compare geometrically?
    \item How do state-of-the-art multiclass debiasing techniques compare considering the discrimination of a downstream application?
    
\end{itemize}

To address which state-of-the-art debiasing technique performs religious debiasing the best, an extensive background on social biases in word embeddings is given. The evaluation metrics this paper uses to access performance are explained, before the debiasing techniques examined are illustrated. This paper, then, highlights the need for religious debiasing by showing its presence in a word embedding. Consequently a common base for the analysis of bias removal is established to compare the debiasing methods. Finally, this paper discusses the performance of the debiasing techniques and based on this evaluation, advises the use of one. 

\section{Background}
\label{Background}
%biases in word embeddings
Social biases have been found in popular, widely used word embeddings such as GloVe \cite{pennington2014glove} or word2Vec \cite{mikolov2013efficient}, \cite{caliskan2017semantics}. 
Specifically, gender biases have been found to persist by creating simple analogies, which have led to the example "Man is to Computer Programmer as Woman is to Homemaker"  \cite{caliskan2017semantics},\cite{bolukbasi2016man}. This analogy clearly shows that the word embeddings have captured gender bias with regards to occupation, which may cause disruption in, e.g.\, a CV-Scanning application. Similarly, the multi-class racial bias in word embeddings has led to other biased analogies \cite{manzini2019black} being coined. Sweeney and Najafan have also shown that multi-class bias based on nationality or religion is present in word embeddings, which endangers specific identity groups to be treated differently \cite{sweeney2019transparent}. 

Social biases have, therefore, been proven to likely exist within word embeddings. As mentioned before (\ref{intro}), biases in data driven artificial intelligence and ,thus, word embeddings have many causes, especially related to the bias present in the data used. Papakyriakopoulos, Hegelich, Serrano, and  Marco find that  biases in word embeddings are closely related to the input training data \cite{papakyriakopoulos2020bias}. In fact, even when the text used for training was written for a "formal and controlled environment like Wikipedia, [it] result[ed] in biased word embeddings"(p.455, \cite{papakyriakopoulos2020bias}). 

A strong cause for bias in textual data is the more frequent co-occurrence of particular words to the identity terminology of one group rather than the other(s). Word embedding algorithms typically take co-occurrences as an indicator of context and semantic relationships. Thus, the word embeddings learn a stronger association between, for example, 'woman' and 'nurse' than 'man' and 'nurse'. This association, however, is an example of a stereotype, which should ideally not be captured in the artificial intelligence applications.
Garg, Schiebinger, Jurafsky and Zou confirm that word embeddings "accurately capture both gender and ethnic occupation percentages" (p.3636, \cite{garg2018word}).

%Binary Debiasing
The biases within word embeddings can amplify through an application, causing unfair results, which may influence actions in the real world. This, in turn, may lead to unequal treatment based on certain sensitive attributes and actively cause discrimination. Hence, it is vital to establish mitigation methods. 

Debiasing methods may tend to different categories of biases. For instance, debiasing binary biases mitigates the unequal treatment of two groups based on a sensitive feature, and joint debiasing mitigates biases based on various sensitive attributes simultaneously.
This paper demonstrates a multi-class debiasing, which deals with bias across more than two groups, by considering three religious groups, namely: Christianity, Islam, Judaism.
%Binary biases are concerned with the unequal treatment of two groups. The most popular and well researched example here is gender bias \cite{caliskan2017semantics}, \cite{bolukbasi2016man}. Binary debiasing may also concern two of many groups, which are established via a multi-class sensitive feature, such as race.
%In contrast to binary debiasing, multi-class debiasing deals with bias across groups larger than two. This paper's running example for this is religion with the groups: Christianity, Islam, Judaism.
%In addition to the consideration of how many identity groups one is debiasing, research has also been performed on joint debiasing. Joint debiasing attempts to mitigate biases based on various sensitive features simultaneously (e.g.\ it attempts to debias against race, religion and gender in one step, rather than the multiple consecutive applications of an original debiasing method).
The development of debiasing techniques is novel research, yet a few state-of-the-art approaches have been proposed. Following the notion that word embedding biases are a direct result of bias in the data, Brunet, Alkalay-Houlihan, Anderson, and Zemel have proposed a technique to track which segment of data is responsible for some bias \cite{brunet2018understanding}. It follows naturally that this can be applied as a debiasing technique by omitting these segments when training the word embedding model. Most debiasing techniques, however, concentrate on post-processing pre-trained word embeddings.

Bolukbasi, Chang, Zou, and Saligrama propose soft and hard debiasing as binary debiasing methods \cite{bolukbasi2016man}, which Manzini, Lim, Tsvetkov, and Black transfer into the multi-class domain \cite{manzini2019black}. Popovic, Lemmerich and Strohmaier expand these debiasing techniques further into SoftWEAT and hardWEAT, which also are applicable for joint debiasing \cite{popovic2020joint}. Another joint multiclass debiasing approach is the Conceptor debiasing method by Karve, Ungar and Sedoc \cite{karve2019conceptor}.
%MUlticlass debiasing (Religion as an example too)

% Critics on Debiasing Methods so far
With the increased research into debiasing methods, Gonen and Goldberg \cite{GONEN19} provide a critical view on the effectiveness of debiasing. The removal of bias in the techniques, such as hard debiasing, relies on the definition of the bias as being the projection onto a biased subspace. Gonen and Goldberg, however, believe that this is a mere indication of the presence of bias. Thus, although the debiasing methods may eliminate the bias projections, the bias is still captured within the geometry of supposedly neutralized words \cite{GONEN19}. Hence, it is important to consider the quantification of bias removal critically. 

In this paper, the multi-class debiasing methods, all mentioned above, namely Hard debiasing, SoftWEAT debiasing and Conceptor debiasing will be evaluated on different metrics in an attempt to quantify bias removal from geometrical and down stream perspectives.
Previous work comparing debiasing techniques have evaluated their performance on merely one geometric metric quantifying bias \cite{bolukbasi2016man}, \cite{manzini2019black},\cite{karve2019conceptor}, whereas this paper uses two geometric metrics, in addition to utilizing a downstream bias metric. 

These metrics and debiasing techniques will now be introduced, before an investigation of religious bias, as an example of multiclass social bias, is conducted on a word embedding. Having established the need for religious debiasing, the bias removal will be conducted and analysed.

\section{Methodology}

\subsection{Terminology}

To aid in the explanation of the debiasing techniques and evaluation metrics, some definitions and terminologies are introduced first. 
\begin{itemize}
\item A class $C$ consists of a set of protected groups defined by some criteria, like religion or race. 
\item A subclass $S_c$ then refers to a particular protected group within that class, such as Judaism when considering the religion class.
\item An equality set $E$ for a class is a set containing a term for each subclass, where all terms can be considered to denote an equivalent concept within each subclass. Thus, for instance, an equality set for $C$ = religion with $S_c$ = (Christianity, Islam, Judaism) could be (Church, Mosque, Synagogue).
\item A target set $T$ is a set of identity terms referring to a  particular sub-class, thus inherently carrying bias. For Christianity this could include: \{Church, Churches, Bible, Bibles, Jesus\}
\item An attribute set $A$ contains sets of words referring to several topics, none of which should, in principle, be linked to the target set of a subclass, but that a target set of words may be associated to \cite{popovic2020joint}. The aim of the debiasing methods is to remove this link. Examples for attribute sets are collections of words considered to be pleasant, or unpleasant, respectively or collections of words describing notions such as families, arts or occupations.
\end{itemize}

\subsection{Bias Measurements Techniques}
\label{evalMeth}
To quantify the bias removal, the three metrics introduced below are used. The first two metrics introduced evaluate the removal geometrically by considering the cosine distance of target and attribute sets, whereas the third highlights bias presence via a simple sentiment analysis application.

\subsubsection{Word Embedding Association Test (WEAT)}
\label{WEAT}

The standard evaluation of bias is the Word Embedding Association Test (\emph{WEAT}) as established by Caliskan, Bryson, and Narayanan. It is widely used, for instance in \cite{bolukbasi2016man} and \cite{popovic2020joint}, and it has been expanded, for instance, to the Sentence Encoder Association Test (SEAT) \cite{may2019measuring}.

%The bias measurement via WEAT utilizes target and attribute sets as defined above by considering a pair of target and attribute sets simultaneously. A statistical test is performed to determine whether a target set of one subclass $T_1$ is related to an attribute set $A_1$ in a similar fashion that a target set of another subclass $T_2$ is related to attribute set $A_2$.

WEAT tests the association between one target and attribute set, relative to the association of the other target and attribute set in order to examine the null hypothesis that both target sets are equally similar to both attribute sets and not exhibiting any bias \cite{caliskan2017semantics}.

To perform WEAT, the mean cosine similarity of the target set $T_1$ to attribute sets $A_1$ and $A_2$ is compared to the mean cosine similarity of the target set $T_2$ to $A_1$ and $A_2$. The exact calculations for the test statistic $S(T_1,T_2,A_1,A_2)$ and the effect size $d$ of the two attribute - target set pairs is given below. 
Let $s(w, A_1, A_2)$ be defined as in equation \ref{eq:s}, where $w$ is a given word vector:
\begin{equation}
\label{eq:s}
s(w,A_1,A_2) = mean_{a_1 \in A_1} cos(\vec{w},\vec{a}_1)- mean_{a_2 \in A_2} cos(\vec{w},\vec{a}_2)
\end{equation}

\begin{equation}
S(T_1,T_2,A_1,A_2) = \sum_{t_1\in T_1} s(t_1,A_1,A_2) - \sum_{t_2\in T_2} s(t_2,A_1,A_2),
\end{equation}
The effect size $d$ quantifies how distant these two associations of target and attribute pairs are. The closer the effect size $d$ is to zero, the less distant the two associations are and thus, the less bias can be found between the target and attribute sets \cite{caliskan2017semantics}. 
\begin{equation}
d = \frac{mean_{t_1 \in T_1} s(t_1,A_1,A_2) - mean_{t_2 \in T_2}s(t_2,A_1,A_2)}{std\text{-}dev_{w \in T_1 \cup T_2} s(w,A_1,A_2)}
\end{equation}
    
It should be noted that bias here is defined on the relative distances.

\subsubsection{Mean Average Cosine Similarity (MAC)}
\label{MAC}
WEAT as proposed by Caliskan et al. \cite{caliskan2017semantics} provides a geometric interpretation of the distance between two sets of target words and two sets of attribute words.

The mean average cosine similarity (\emph{MAC}) uses the intuition behind WEAT and applies this notion to a multiclass domain as proposed by Manzini et al. \cite{manzini2019black}. Instead of comparing the associations of one target set $T_1$ and an attribute set $A_1$, to the association of $T_2$ and $A_2$, MAC considers the association of one target set $T_1$ to all attribute sets $A$ at one time. 

The MAC metric is computed by calculating the mean over the cosine distances between an element $t$ in a target set $T$ to each element in an attribute set $A$, as seen in equation \ref{eq:SMAC}, in which the cosine distance is defined as $ cos_{distance}(t,a) = 1 - cos(t,a)$.  This is repeated for all elements in $T$ to all attribute sets. The MAC then describes the average cosine distance between each target set and all attribute sets.
\begin{equation}
\label{eq:SMAC}
s_{MAC}(t,A_j) = \frac{1}{|A_j|} \sum_{a \in A_j} cos_{distance}(t,a)
\end{equation}

%Since the cosine distance is defined here as $ cos_{distance}(t,a) = 1 - cos(t,a)$, the cosine distance ranges from 0 to 2: with 0 indicating that the vectors are the same and 2 indicating that the vectors have maximum cosine distance. Thus, to interpret the MAC metric, one should expect a MAC value of close to 1 with unbiased word embeddings. This would indicate that the target sets are as close to an attribute set, as they are distant from it.

\subsubsection{Relative Negative Sentiment Bias (RNSB)}
\label{RNSB}
The relative negative sentiment bias (\emph{RNSB}) is an approach proposed by Sweeney and Najafan \cite{sweeney2019transparent} in order to offer insights on the effect of biased word embeddings through downstream applications. Its framework involves training a logistic classifier to predict the positive or negative sentiment of a given word. The classifier is trained on supposedly unbiased sentiment words, which are encoded via the word embedding to be investigated. Sweeney and Najafan then encode identity terms and predict their respective negative sentiment probability. These results are used to form a probability distribution $P$. 
Intuitively, unbiased word embeddings would result in this probability distribution to be uniform, i.e. each class has equal probability of being classified as of negative sentiment.
The RNSB is then defined as Kullback-Leibler divergence of $P$ from the uniform distribution $U$ \cite{sweeney2019transparent}.

\subsection{Debiasing Techniques}

These three metrics will be used to quantify the bias removal in the three debiasing techniques considered in this paper. Namely, these are Hard debaising, SoftWEAT and Conceptor debiasing.

\subsubsection{Hard Debiasing}
Bolukbasi et al. \cite{bolukbasi2016man} established two binary debiasing methods, namely: Soft and Hard debiasing, which Manizini et al. \cite{manzini2019black} then applied to the multiclass domain. These approaches mainly rely on two steps: The identification of a bias subspace, and the subsequent removal of that bias. The main difference between these two methods is the severity of bias removal.%: Hard debiasing forces neutral words to zero in the bias subspace, whereas soft debiasing dampens the bias subspace components \cite{bolukbasi2016man}.

The bias subspace identification utilizes equality sets $E_i$. For each set, the center of the set is computed and the distance of each term in the equality set to the center is considered. The subspace capturing the class is then found by examining the variance of each term. Bias removal is carried out by a `neutralize and equalize' approach. The projection of words that are declared neutral onto the bias subspace is subtracted from their word vector. The identity words, however, rely on their bias component. Thus, in the equalization step, the terms within an equality set, are centralized and are each given an equal bias component.

\subsubsection{SoftWEAT Debiasing}
\label{SW}
Popovic et al. propose debiasing techniques SoftWEAT and hardWEAT \cite{popovic2020joint}, which borrow intuition from WEAT \cite{caliskan2017semantics}.
SoftWEAT expands the target set of each subclass by considering the $n$ closest neighbours to all identity terms. Merely this set is then manipulated. To find the linear transformation to be applied, the attribute sets the target set of a subclass is biased against is found via WEAT and their respective null space vectors are calculated. The translation of the subclass embeddings is then taken from the null space vector, which decreases the WEAT score the most. The final transformation can be scaled by hyper-parameter $\lambda$.

\subsubsection{Conceptor Debiasing}
Karve et al. developed the Conceptor debiasing post processing method \cite{karve2019conceptor}. The notion of this method is to generate a conceptor, as defined by Jaeger \cite{jaeger2014controlling}, to represent bias directions and to subsequently project these biased directions out of the word embeddings. 

A square matrix conceptor $C$ is a regularized identity map, which maps an input to another -- in the debiasing domain, a word embedding to its bias \cite{karve2019conceptor}. For the exact mathematical definition of a conceptor readers can refer to \cite{liu2019unsupervised} and \cite{karve2019conceptor}. % This paper will briefly state the mathematical definition of a conceptor as given by Liu, Ungar and Sedoc's and Karve et al.'s work \cite{liu2019unsupervised}, \cite{karve2019conceptor}. For an elaborate exploration of conceptors it is referred to Jaeger's work \cite{jaeger2014controlling}, \cite{jaeger2014conceptors}.
%The conceptor matrix $C$ attempts to minimize equation \ref{eq:conc}, where $Z$ is the union of all subclass target sets for a class, the scalar $\alpha$ is the aperture and $||.||_{F}$ is calculated as the Frobenius norm. The aperture constrains the conceptor to be the zero mapping at zero aperture and lets the conceptor approach identity mapping, as the aperture approaches infinity \cite{jaeger2014conceptors}. Hence, the aperture controls the focus of the conceptor.
%\begin{equation}
%\label{eq:conc}
%||Z -CZ||_{F}^2+\alpha^{-2}||C||_{F}^2
%\end{equation}
%The conceptor matrix minimizing equation \ref{eq:conc} is found via the closed form solution derived from equation \ref{eq:conc2}, where $n$ is the number of words defined in $Z$.
%\begin{equation}
%\label{eq:conc2}
%C = \frac{1}{n}ZZ^T(\frac{1}{n}ZZ^T+\alpha^{-2}I)^{-1}
%\end{equation}
%Its basis is defined by the principal components of the patterns it is learnt from. Thus if the conceptor is learnt on the identity terms of 
%The characteristics of conceptors that lend themselves for such a domain application are mainly twofold: 
Conceptors can be manipulated through boolean logic. Thus, to project out a bias subspace, one can apply the negated conceptor (representing the bias directions) to the word embeddings. In addition to this, through the use of boolean logic, multiple conceptors generated for various class biases can be combined, enabling joint debiasing \cite{karve2019conceptor}. Moreover, a conceptor provides a soft projection \cite{jaeger2014controlling}. For debiasing this means, that the conceptor dampens the bias directions captured in it. Hence, the soft projection will alter only some components of some embeddings, leaving others largely unaltered \cite{jaeger2014conceptors}. 

\section{Analysis of Religious Bias in Word Embeddings}
\label{AN}

\subsection{Data}
\label{data}

Each of the debiasing approaches described is based on different types of data: Conceptor debiasing utilizes a set of unlabeled biased words, Hard debiasing requires equality sets, and SoftWEAT is based on the target and attribute sets of WEAT. This paper will attempt to debias against the religion class, specifically with the subclasses: Christianity, Islam, Judaism. The equality set used for religious multiclass debiasing in Manizini et al.'s paper \cite{manzini2019black} is extended by hand to include 11 equality sets, which are available for downloading\footnote{https://github.com/thaleaschlender/An-Evaluation-of-Multiclass-Debiasing-Methods-on-Word-Embeddings}. The attribute sets used in this paper are inspired from Popovic et al.'s work \cite{popovic2020joint}. 

Finally, the debiasing methods are applied on three established word embedding representations, namely: Word2Vec\footnote{https://code.google.com/archive/p/word2vec/}, GloVe\footnote{https://nlp.stanford.edu/projects/glove/} and ConceptNet \footnote{http://blog.conceptnet.io/posts/2019/conceptnet-numberbatch-19-08/}.

\subsection{Analysis}

Social biases are present in the word embeddings when neutral words are more strongly associated with one subclass than another. In this section it is shown what impact these associations have more specifically to each subclass of religion: Christianity, Islam, and Judaism.

In order to quantify captured stereotypes in word embeddings, analogies are scored, as proposed by Bolukbasi et al. \cite{bolukbasi2016man}. %The analogies take on the simple form: " \textit{a} is to \textit{x}, as \textit{b} is to \textit{y}", an example being " \textit{kitten} is to \textit{cat}, as \textit{puppy} is to \textit{dog}". 
The analogies are then scored via equation \eqref{eq:analo}, where $\delta$ is the similarity threshold and $\vec{a}, \vec{b},\vec{x},\vec{y}$ are words as given above. The intuition behind this equation is that an analogy capturing relationships well should have directions $\vec{a}-\vec{b}$ and $\vec{x}-\vec{y}$ approach parallelism. 

\begin{equation}
\label{eq:analo}
S_{(a,b)}(x,y) = 
\begin{cases}
    cos(\vec{a}-\vec{b},\vec{x}-\vec{y}) & \text{if } ||\vec{x}-\vec{y}|| \leq \delta\\
    0,              & \text{otherwise}
\end{cases}
\end{equation}

Table \ref{analo_tab} lists the analogies with a score of over 0.15, that are established within the word2Vec embeddings. As a comparison, the biased analogy established by Bolukbasi et al. \cite{bolukbasi2016man} and Manzini et al. \cite{manzini2019black}, in addition to some appropriate analogies, are given with their respective scores. Although it follows that the maximal absolute score of equation \eqref{eq:analo} is 1, in table \ref{analo_tab} one can see that established analogies like "\textit{kitten} is to \textit{cat}, as \textit{puppy} is to \textit{dog}", achieve a score of 0.38. Thus, when regarding how high appropriate analogies are scored, biased analogies with an absolute score of higher than 0.15 indicate that these biased analgoies are captured in the word embeddings.

An appropriate analogy concerning religion would be "\textit{Muslim} is to \textit{Islam} as \textit{Christian} is to \textit{Christianity}", which describes the correct correspondence of religion and its members. However, a similarly high classified analogy is "\textit{Christian} is to \textit{judgemental} as \textit{Muslim} is  to \textit{terrorist}". This wrong association of religions to terrorist and judgmental is an unjust example of a captured stereotype in the word embedding. The prejudice of Muslims being more strongly associated with violence and terrorism is deeply embedded in society as proven by Sides and Gross. They hypothesize and confirm that "Americans will stereotype Muslims negatively on the warmth dimension— that is, as threatening, violent, etc" (p.5, \cite{sides2013stereotypes}).

\begin{table}[h]
\caption{Analogies scoring higher than .15 in Word2Vec}
\begin{center}
\label{analo_tab}
\scalebox{0.75}{
%from 1.15 upwards
\begin{tabular}{|c|c|}
\hline
\textbf{\textsc{Analogy}} & \textbf{\textsc{score}}\\
\hline
\multicolumn{2}{|c|}{\textbf{Appropriate Analogies}}\\
\hline
\textit{cat} is to \textit{kitten} as \textit{dog} is  to \textit{puppy} & .38332\\
\textit{Muslim} is to \textit{Islam} as \textit{Christian} is to \textit{Christianity}& .27088\\
\textit{Christian} is to \textit{Christianity} as \textit{Jew} is to \textit{Judaism}& .26884\\
\textit{Muslim} is to \textit{Islam} as \textit{Jew} is to \textit{Judaism}&.24883\\
\textit{Christianity} is to \textit{Church} as \textit{Judaism} is to \textit{Synagogue} & .24054\\
\hline
\multicolumn{2}{|c|}{\textbf{Analogies Exhibiting Stereotypes}}\\
\hline
\textit{woman} is to \textit{homemaker} as \textit{man} is  to \textit{programmer} & .26415\\
\textit{Black} is to \textit{criminal} as \textit{Caucasian} is  to \textit{police} &  .07325\\
\hline
\textit{Christian} is to \textit{judgemental} as \textit{Muslim} is  to \textit{terrorist} & .246935 \\
\textit{Christian} is to \textit{conservative} as \textit{Muslim} is  to \textit{terrorist} & .215955\\
\textit{Christian} is to \textit{conservative} as \textit{Muslim} is  to \textit{liberal} & .177172\\
\textit{Christian} is to \textit{judgmental} as \textit{Muslim} is  to \textit{uneducated} & .171767\\
\textit{Christian} is to \textit{judgmental} as \textit{Muslim} is  to \textit{violent} & .171105\\
\textit{Christian} is to \textit{greedy} as \textit{Muslim} is  to \textit{terrorist} & .166391\\
\textit{Christian} is to \textit{judgmental} as \textit{Muslim} is  to \textit{liberal} & .155485\\
\hline
\textit{Jew} is to \textit{hairy} as \textit{Christian} is  to \textit{conservative} & .222206\\
\textit{Jew} is to \textit{greedy} as \textit{Christian} is  to \textit{conservative} & .213083\\
\textit{Jew} is to \textit{greedy} as \textit{Christian} is  to \textit{judgmental} & .201595\\
\textit{Jew} is to \textit{hairy} as \textit{Christian} is  to \textit{judgmental} & .197683\\
\textit{Jew} is to \textit{liberal} as \textit{Christian} is  to \textit{conservative} & .181528\\
\textit{Jew} is to \textit{cheap} as \textit{Christian} is  to \textit{conservative} & .177668\\
\textit{Jew} is to \textit{dirty} as \textit{Christian} is  to \textit{conservative} & .176638\\
\textit{Jew} is to \textit{familial} as \textit{Christian} is  to \textit{conservative} & .173743\\
\textit{Jew} is to \textit{hairy} as \textit{Christian} is  to \textit{violent} & .168193\\
\textit{Jew} is to \textit{dirty} as \textit{Christian} is  to \textit{judgmental} & .151427\\
\hline
\textit{Muslim} is  to \textit{terrorist} as \textit{Jew} is to \textit{greedy}   & .239060\\
\textit{Muslim} is  to \textit{terrorist} as \textit{Jew} is to \textit{hairy}   & .227352\\
\textit{Muslim} is  to \textit{violent} as \textit{Jew} is to \textit{greedy}  & .207468\\
\textit{Muslim} is  to \textit{violent} as \textit{Jew} is to \textit{hairy}  & .196129\\
\textit{Muslim} is  to \textit{terrorist}as \textit{Jew} is to \textit{dirty}   & .192120\\
\textit{Muslim} is  to \textit{terrorist} as \textit{Jew} is to \textit{cheap}  & .187418\\
\textit{Muslim} is  to \textit{uneducated} as \textit{Jew} is to \textit{greedy}   & .180224\\
\textit{Muslim} is  to \textit{conservative} as \textit{Jew} is to \textit{greedy}   & .172667\\
\textit{Muslim} is  to \textit{terrorist} as \textit{Jew} is to \textit{familial}   & .168889\\
\textit{Muslim} is  to \textit{liberal} as \textit{Jew} is to \textit{greedy}   & .160143\\
\textit{Muslim} is  to \textit{violent} as \textit{Jew} is to \textit{dirty}   & .155248\\
\textit{Muslim} is  to \textit{conservative} as \textit{Jew} is to \textit{hairy}  & .154570\\
\hline
\end{tabular}}
\end{center}
\end{table}

%. In what ways are these system behaviors harmful, to whom are they harmful, and why?

%. What are the social values (obvious or not) that underpin this conceptualization of “bias?”

\section{Experiments and Results}
\label{exp}

\subsection{Experimental Setup}
After the confirmation of religious bias existence two main sets of experiments are held and described below.

The first aims to evaluate the performance of bias removal techniques on a common basis. It does this by observing different quantifications of bias pre- and post- the application of the debiasing methods. The metrics RNSB, WEAT and MAC are calculated for each word embedding, Word2Vec, GloVe and ConceptNet. We use hard debiasing, Conceptor debiasing with the aperture $\alpha = 10$ and SoftWEAT with $\lambda = 0.5$ and a threshold of 0.5. After each debiasing method, the metrics are calculated anew. Thus, it is possible to evaluate the performance of prior and post debiasing on different word embeddings and debiasing methods in a universal, comparable manner. Since WEAT and MAC are distance measures, the results collected here remain stable over multiple runs. However, to calculate the RNSB metric a logistic classifier is trained on randomly split training and test data. Hence, variability in the RNSB metric is introduced through the individually trained classifier. To counteract this, the RNSB is averaged over 20 runs.

Afterwards, a second set of experiments aims to examine the impact of the SoftWEAT hyperparameters by investigating the impact of hyperparameter $\lambda$. This parameter tunes how harshly debiasing is applied and is named as one of the strong advantages of SoftWEAT \cite{popovic2020joint}.

\subsection{RNSB Metric on Word Embeddings}

%RNSB Results
The results in table \ref{RNSB_tab} show the RNSB values before and after hard debiasing, Conceptor debiasing and SoftWEAT debiasing approaches on word2Vec, GloVe and ConceptNet respectively. The best RNSB scores of each word embedding is highlighted. To statistically analyse whether the RNSB has been improved significantly, a one tailed t-test is performed on all values. The $p$ values are given in table \ref{RNSB_tab} showing that with a significance of $\alpha = 0.05$, it can be concluded that each debiasing method improves the mean RNSB value significantly compared to the non-debiased word embeddings.

Pre-debiasing the word embeddings of ConceptNet carry the least bias, whereas the GloVe word embeddings carry the most bias, according to their RNSB score.
Hard debiasing appears to debias the embeddings most efficiently, followed by Conceptor debiasing, whereas SoftWEAT achieves worse results in comparison. This could be attributed to the fact that SoftWEAT only manipulates a collection of words (the identity terminology and its neighbours), whereas the other two debiasing approaches manipulate the whole vocabulary.

\begin{table}[htbp]
\caption{Relative Negative Sentiment Bias after application of debiasing techniques on Word2Vec, GloVe and ConceptNet}
\begin{center}
\label{RNSB_tab}
\begin{tabular}{|c|c|c|c|c|c|c|}
\hline
\textbf{Debiasing}&\multicolumn{6}{|c|}{\textbf{Word Embeddings}} \\
\cline{2-7} 
\textbf{Techniques} &\multicolumn{2}{|c|}{\textbf{\textit{Word2Vec}}}& \multicolumn{2}{|c|}{\textbf{\textit{GloVe}}}&\multicolumn{2}{|c|} {\textbf{\textit{ConceptNet}}}\\
\cline{2-7}
& \textit{RNSB} & \textit{p} & \textit{RNSB} & \textit{p} & \textit{RNSB} & \textit{p} \\
\hline
Non-Deb. & 0.12339 & N/A & 0.26033 & N/A &0.02276 & N/A \\ 
Conc. Deb. & 0.00682 & 0.027 & 0.00024 & 0.002 &0.00775 & 0.031 \\
Hard Deb. & \textbf{0.0} & 0.017 & \textbf{0.00023}& 0.002 &\textbf{0.0} & 0.024\\
SoftWEAT & 0.07244 & 0.032 & 0.0525& 0.002 &0.0179 & 0.035\\
\hline
\end{tabular}
\end{center}
\end{table}

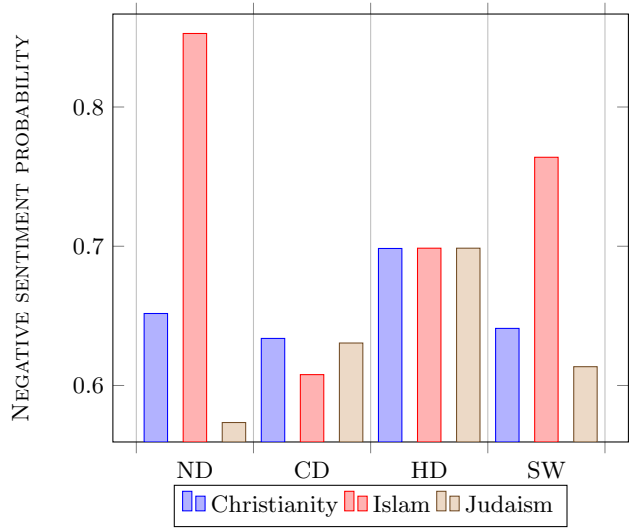
\begin{figure}[h]
    \centering
\begin{tikzpicture}
\title{\textsc{RNSB}}
\begin{axis}[
	x tick label style={
		/pgf/number format/1000 sep=},
    enlarge x limits={abs=2cm},
    symbolic x coords={ND, CD, HD, SW, DUMMY},
	ylabel= \textsc{Negative sentiment probability},
	enlargelimits=0.05,
	legend style={at={(0.5,-0.1)},
	anchor=north,legend columns=-1},
	ybar interval=0.6,
	xtick={ND, CD, HD, SW, DUMMY},
]
%SUBTRACT FROM 1 OR FROM BASELINE
\addplot % CHRISTIANITY
coordinates{(ND, 0.6517) (CD,0.6338)  (HD, 0.6984) (SW, 0.6410) (DUMMY, 0.7)};
\addplot % CD
coordinates{(ND,0.8528 ) (CD, 0.6078)  (HD, 0.6986) (SW,  0.7639)(DUMMY, 0.7)};
\addplot % HD
coordinates{(ND,0.5734 ) (CD, 0.6305)  (HD, 0.6986) (SW, 0.6135)(DUMMY, 0.7)};
%HORIZONTAL LINE AT ONE
\legend{Christianity, Islam, Judaism}
%Relative improvement
\end{axis}
\end{tikzpicture}
\caption{The negative sentiment probability for Religion terminology from Christianity, Islam and Judaism before and after post processing methods, namely: ND: no debiasing, CD: Conceptor debiasing, HD: hard debiasing and SW: SoftWEAT debiasing}
\label{fig:RNSBBAR}
\end{figure}

The RNSB metric aims to evaluate the bias through a downstream sentiment analysis task. The results show that post debiasing each religion is classified more equally negative with respect to the other religions. Concretely, these improvements for the three debiasing methods on Word2Vec can be seen in figure \ref{fig:RNSBBAR}, which depicts the negative sentiment probability for each religion.

The RNSB score decreases as the negative sentiment probability for each religion approaches a sample of the uniform distribution. In figure \ref{fig:RNSBBAR}, one can compare each distribution to a fair uniform distribution. Observing this, the non debiased distribution differs from the uniform distribution considerably, whereas the post hard debiasing distribution resembles the uniform distribution the most. 
This is also indicated by their respective RNSB scores shown in table \ref{RNSB_tab}. 

Furthermore, figure \ref{fig:RNSBBAR} shows that Islam terminology is most likely to be predicted as of negative sentiment. This considerable difference is intuitive when recalling the Muslim and terrorism association captured in the word2Vec embedding, found in the analogies of table \ref{analo_tab}. It is also interesting to note that after performing Conceptor debiasing, Islam terminology actually becomes the least likely to be predicted of negative sentiment. Thus, Conceptor debiasing has changed the hierarchy of the religions, whereas hard debiasing and SoftWEAT debiasing dampen the original non-debiased distribution.

\subsection{WEAT and MAC on Word Embeddings}
This paper now moves on from the downstream application analysis via RNSB to the geometric analysis of the bias removal methods via WEAT and MAC. Again, to identify the impact of each debiasing method, all values can be compared to the original word embedding prior to any debiasing. 

%WEAT Results
Firstly, the WEAT measurements prior and post the three debiasing methods are shown in table \ref{WEAT_tab}. To ease the interpretation of the table, the best scores are bold, whilst scores, which decrease performance to the baseline of the non debiased word embeddings are italic. With the exception of the SoftWEAT application on the ConceptNet embedding, all debiasing methods reduce the WEAT measurements and thus, appear to debias the word embeddings to a given extent. 

The performance of the three debiasing techniques in terms of WEAT scores is the same as found within the RNSB evaluation. The hard Debiasing technique performs best, followed by Conceptor debiasing, whereas SoftWEAT's WEAT scores are poor in comparison. In fact, when applying SoftWEAT to ConceptNet, it actually increases the WEAT score, indicating an increase of measured bias. This poor performance could be attributed to the manipulation of less of the embeddings in the vocabulary, as mentioned earlier.

\begin{table*}[htbp]
\caption{WEAT and $|$1-MAC$|$ after application of debiasing techniques on word2Vec, GloVe and conceptnet - The closer to 0 the better}
\begin{center}
\label{WEAT_tab}
\label{MAC_tab}
\begin{tabular}{|c|c|c|c|c|c|c|}
\hline
&\multicolumn{6}{|c|}{\textbf{Word Embeddings}} \\
\cline{2-7} 
 \textbf{Debiasing}&\multicolumn{3}{|c|}{\textbf{WEAT scores}} & \multicolumn{3}{|c|}{\textbf{$|$1-MAC$|$}} \\
\cline{2-7} 
\textbf{Techniques}&\textbf{\textit{Word2Vec}}& \textbf{\textit{GloVe}}& \textbf{\textit{ConceptNet}}& \textbf{\textit{Word2Vec}}& \textbf{\textit{GloVe}}& \textbf{\textit{ConceptNet}} \\
\hline
Non-Debiased & 0.39469 & 0.67556 & 0.76714 & 0.11787 & 0.16771 & 0.00482\\
Conceptor Debias & 0.17112 & 0.06348 & 0.30251& \textbf{0.00436} & \textbf{0.0003} & \textbf{0.0030}\\
Hard Debias & \textbf{0.00082} & \textbf{0.038215} & \textbf{0.00441} & 0.11039 & 0.15603 & \textit{0.00624}\\
SoftWEAT & 0.31639 & 0.40967 & \textit{0.83589} & 0.07766 & 0.11871& \textit{0.01367}\\

\hline
\end{tabular}
\end{center}
\end{table*}

%MAC Results

In table \ref{MAC_tab} the MAC scores are presented. In order to ease comparison, the MAC values are subtracted from the optimal value 1. Hence, the closer the MAC values are to 0, the less bias was measured. A similar performance hierarchy of debiasing techniques found in RNSB and WEAT is expected for the MAC scores. Again, to ease comparison, bold and italic fonts are used as described above.

Via the one tailed t-test, the corresponding $p$ values to the MAC scores were calculated. With a significance of $\alpha = 0.01$, the MAC values are all improved compared to their non-debiased version, an exception being both SoftWEAT and hard debiasing when applied to ConceptNet.

Both WEAT and MAC are taken from the notion of measuring bias in cosine distance. The results of both metrics show that the Conceptor debiasing performs well, whilst SoftWEAT performs poorly in comparison. It is interesting to note that hard debiasing achieves the best RNSB and WEAT scores, yet achieves poor MAC scores - worsening the MAC score within the ConceptNet embeddings. This could be due to the fact that WEAT is a relative measure between two religions and two attribute sets, whereas MAC captures the distance of one religion to all attribute sets. Hard debiasing may introduce new bias by the harsh removal of its religion subspace. This bias introduction may then only be captured in the MAC scores. In fact, when examining the measured mean cosine distance for each religion to each attribute set in word2Vec, one can see that Hard Debiasing improves scores for Judaism, but slightly worsens scores for Christianity and Islam.

In general the results above show that the word embedding ConceptNet carries the least bias as evaluated by MAC and RNSB scores. However, surprisingly, the WEAT score measured in ConceptNet is the worst of all three. The GloVe embeddings seem to carry the most bias concerning the RNSB and MAC metrics, which is intuitive when considering the common crawl data it was trained on.

\subsection{SoftWEAT hyperparameter $\lambda$ experimentation}

Having analysed the general performance of all three debiasing techniques above, this paper now turns to the evaluation of SoftWEAT, which has performed most poorly so far. The analysis will examine whether the tuning of the hyperparameter $\lambda$ may improve the performance within the evaluation metrics used above. %The set up used has been established in section \ref{SWexp}. %A similar investigation is done for the threshold hyperparameter and can be found in the appendix \ref{thresh}. The results of the hyperparameter $\lambda$ experiments can be seen in the graphs below. 

In figure \ref{fig:SWWEAT} it can be seen that the WEAT score monotonically decreases with increasing values up to a $\lambda$ of 0.6. From then onwards, the WEAT score steadily increases again. Popovic et a.l \cite{popovic2020joint} report a similar peek in their religious debiasing of Word2Vec. It seems that with a $\lambda$ higher than 0.6, new bias is introduced by removing one bias too harshly. However, when regarding the $|$1-MAC$|$ scores in figure \ref{fig:SWMAC}, one can see that higher $\lambda$ values perform better. 

When observing the RNSB scores in figure \ref{fig:SWRNSB}, the tendency that higher $\lambda$ values lead to a general increase in the RNSB score is shown. One should note, however, that the absolute increase between the values is in the small range of 0.031. The variability of the RNSB framework introduced by its anew training of a classifier at each run in addition to the small range of absolute change in the experiments explains the variability in figure \ref{fig:SWRNSB}. Figure \ref{fig:SWRNSB} shows that a good result is already achieved at $\lambda = 0$. This indicates that the RNSB classifications already benefit from the identity terminology of a religion and its neigbours being normalised.

To summarize, it seems that larger $\lambda$ values improve the bias removal in terms of MAC scores, that a peak value is found in the WEAT scores and that the RNSB scores worsen marginally with higher $\lambda$s. %For religous bias removal, we recommend a $\lamda$

%To summarize the findings of the impact in varying $\lambda$, it seems that  To remove religious bias from the word2Vec word embeddings, this paper recommends a $\lambda$ value between 0.6 and 0.8. However, even at these values, SoftWEAT performs the worst in comparison of the RNSB scores to the other approaches in table \ref{RNSB_tab} and WEAT scores to the other approaches in table \ref{WEAT_tab}.
%As mentioned in the discussion of these experiments, the poor SoftWEAT performance may be caused by the fact that the algorithm merely manipulates the word vectors of identity terminology and its neighbours.

\begin{figure}

\begin{subfigure}[b]{0.3\textwidth}  
\centering
\resizebox{\textwidth}{!}{ 
\begin{tikzpicture}
\begin{axis}[
	x tick label style={
		/pgf/number format/1000 sep=},
    enlarge x limits={abs=2cm},
    symbolic x coords={0,0.1,0.2,0.3,0.4,0.5,0.6,0.7,0.8,0.9,1 },
	ylabel= WEAT Score,
	xlabel = SoftWEAT $\lambda$,
	enlargelimits=0.05,
	legend style={at={(0.5,-0.1)},
	anchor=north,legend columns=-1},
]
\addplot[
    color=blue,
    mark=square,
    ]
coordinates{(0,0.39469)(0.1, 0.37711) (0.2,0.36305)(0.3,0.34791)(0.4,0.33221)
            (0.5,0.31639) (0.6,0.30981) (0.7,0.31566) (0.8,0.32825) (0.9,0.34258) (1,0.35675)};

\end{axis}
\end{tikzpicture}
}
\caption{WEAT score for $\lambda$  values in the range of [0,1], with a threshold of 0.5.}
\label{fig:SWWEAT}
\end{subfigure}
\begin{subfigure}[b]{0.3\textwidth}  
    \centering
\resizebox{\textwidth}{!}{ 
\begin{tikzpicture}

\begin{axis}[
	x tick label style={
		/pgf/number format/1000 sep=},
    enlarge x limits={abs=2cm},
    symbolic x coords={0,0.1,0.2,0.3,0.4,0.5,0.6,0.7,0.8,0.9,1 },
	ylabel=$|$1-MAC$|$ Score,
	xlabel = SoftWEAT $\lambda$,
	enlargelimits=0.05,
	legend style={at={(0.5,-0.1)},
	anchor=north,legend columns=-1}
]

\addplot[
    color=blue,
    mark=square,
    ]
coordinates{(0,0.11788)(0.1, 0.11183) (0.2,0.10477)(0.3,0.09668)(0.4,0.0876)
            (0.5,0.07766) (0.6,0.06704) (0.7,0.05595) (0.8,0.04459) (0.9,0.03314) (1,0.02177)};

\end{axis}
\end{tikzpicture}
}
\caption{$|$ 1 - MAC $|$ score for $\lambda$  values in the range of [0,1], with a threshold of 0.5.}
\label{fig:SWMAC}
\end{subfigure}
\begin{subfigure}[b]{0.3\textwidth}  
\centering
\resizebox{\textwidth}{!}{ 
\begin{tikzpicture}
\begin{axis}[
	x tick label style={
		/pgf/number format/1000 sep=},
    enlarge x limits={abs=2cm},
    symbolic x coords={0,0.1,0.2,0.3,0.4,0.5,0.6,0.7,0.8,0.9,1 },
	ylabel=RNSB Score,
	xlabel = SoftWEAT $\lambda$,
	enlargelimits=0.05,
	legend style={at={(0.5,-0.1)},
	anchor=north,legend columns=-1},
]
\addplot[
    color=blue,
    mark=square,
    ]
coordinates{(0,0.04806)(0.1, 0.05260) (0.2,0.05782)(0.3,0.06129)(0.4,0.06875)
            (0.5,0.07374) (0.6,0.07523) (0.7,0.07790) (0.8,0.08018) (0.9,0.08277) (1,0.08017)};
\end{axis}
\end{tikzpicture}
}
\caption{RNSB score for $\lambda$  values in the range of [0,1], with a threshold of 0.5.}
\label{fig:SWRNSB}
\end{subfigure}
\end{figure}

\section{Conclusion}

This paper analysed the debiasing methods of word embeddings via multiple metrics to establish whether a debiasing method could remove religious bias present in the embeddings.
For this, this paper has reviewed work showing that social biases persist in word embeddings, whilst briefly showing some possible causes in the data word embeddings are trained on. The investigation of state-of-the-art multiclass debiasing methods is done on Hard debaising, SoftWEAT debiasing and Conceptor Debiasing. This paper evaluates their performance not only on the established WEAT metric but also contributes a performance evaluation on the geometric metric MAC and the downstream metric RNSB. By establishing a common base for the debiasing methods, this paper achieves a more meaningful comparison across methods. To highlight the need of the bias removal, religious bias - as an example of social bias - has been shown to persist in word embeddings by scoring various stereotypical analogies.

It is found that Conceptor Debiasing performs well across all metrics and word embeddings, whereas SoftWEAT, regardless of hyperparameter tuning, performs poorly in comparison. Hard debiasing performs well on RNSB and WEAT scores, however shows shortages when evaluating the removal via MAC - indicating that bias may not be removed as well as previously thought. Hence, to recommend a debiasing technique, which performs well in all bias removal quantifications, Conceptor Debiasing is advised. This comes with the added benefit that this technique is applicable for joint multi-class debiasing and is most flexible in what data it is given to establish its conceptor on. % and most flexible

Finally, this paper calls for more research into establishing a common debiasing approach. 
Specifically, this approach should perform well in geometric and downstream analysis of bias removal, whilst not decreasing its semantic power. A possible solution could be a combination of a post processing method as investigated in this paper, with a potential pre selection of data to train on to combat implicit bias.

\bibliographystyle{splncs04}
\bibliography{thyalgorithm}

\end{document}